\documentclass[journal]{IEEEtran}

\usepackage{cite}
\usepackage{amsmath,amssymb,amsfonts}

\usepackage{algorithmic}
\usepackage{algorithm}
\usepackage{graphicx}
\usepackage{textcomp}
\usepackage{subcaption}
\usepackage{textcomp}
\usepackage{xcolor}
\usepackage{mathtools}
\usepackage{makecell}
\usepackage{dsfont}
\usepackage{hyperref}
\usepackage{comment}

\usepackage{array}
\newcolumntype{P}[1]{>{\centering\arraybackslash}p{#1}}

\def\etal{\emph{~et~al. }}
\makeatother

\ifCLASSINFOpdf
\else
\fi

\begin{document}
\title{Prediction-aware and Reinforcement Learning based Altruistic Cooperative Driving}

\author{Rodolfo Valiente$^{1}$, Mahdi Razzaghpour$^{1}$, Behrad Toghi$^{1}$, Ghayoor Shah$^{1}$,  Yaser P. Fallah$^{1}$
\thanks{$^{1}$ Connected \& Autonomous Vehicle Research Lab (CAVREL), University of Central Florida, Orlando, FL, USA. \tt\small {rvalienter90@knights.ucf.edu}}%
}%

\maketitle
\begin{abstract}
\label{sec:abstract}
Autonomous vehicle (AV) navigation in the presence of Human-driven vehicles (HVs) is challenging, as HVs continuously update their policies in response to AVs. In order to navigate safely in the presence of complex AV-HV social interactions, the AVs must learn to predict these changes. Humans are capable of navigating such challenging social interaction settings because of their intrinsic knowledge about other agents’ behaviors and use that to forecast what might happen in the future. Inspired by humans, we provide our AVs the capability of anticipating future states and leveraging prediction in a cooperative reinforcement learning (RL) decision-making framework, to improve safety and robustness.
In this paper, we propose an integration of two essential and earlier-presented components of AVs: social navigation and prediction. We formulate the AV's decision-making process as a RL problem and seek to obtain optimal policies that produce socially beneficial results utilizing a prediction-aware planning and social-aware optimization RL framework. We also propose a Hybrid Predictive Network (HPN) that anticipates future observations. The HPN is used in a multi-step prediction chain to compute a window of predicted future observations to be used by the value function network (VFN). Finally, a safe VFN is trained to optimize a social utility using a sequence of previous and predicted observations, and a safety prioritizer is used to leverage the interpretable kinematic predictions to mask the unsafe actions, constraining the RL policy. We compare our prediction-aware AV to state-of-the-art solutions and demonstrate performance improvements in terms of efficiency and safety in multiple simulated scenarios.

\end{abstract}
\IEEEpeerreviewmaketitle
\section{Introduction}
\label{sec:introduction}
\IEEEPARstart{T}{he} adoption of connected and autonomous vehicles (CAVs) is expected to improve safety and efficiency, decrease traffic accidents, and increase mobility~\cite{cosgun2017towards}. A necessary step toward the widespread integration of autonomous vehicles (AV) in society is allowing coexistence of safe AVs and human-driven vehicles (HVs). 
Nevertheless, coordination and cooperation with HVs are still challenging  for AVs, particularly in complex social interactions~\cite{cosgun2017towards, schwarting2019social, sagberg2015review}. In order to experience those benefits and allow real adoption of AVs on the road, AVs should not only perceive and understand the current environment state but also proactively predict their future states and learn to coordinate and influence other agents. The innate capability to anticipate agents' behaviors and use this knowledge to forecast potential future outcomes allows humans to navigate through complex scenarios; therefore, prediction capabilities are a crucial component in creating secure AVs that can be integrated into society.~\cite{salzmann2020trajectron++, mozaffari2020deep}.

\begin{figure}[t!]
  \centering
  \includegraphics[width=.48\textwidth]{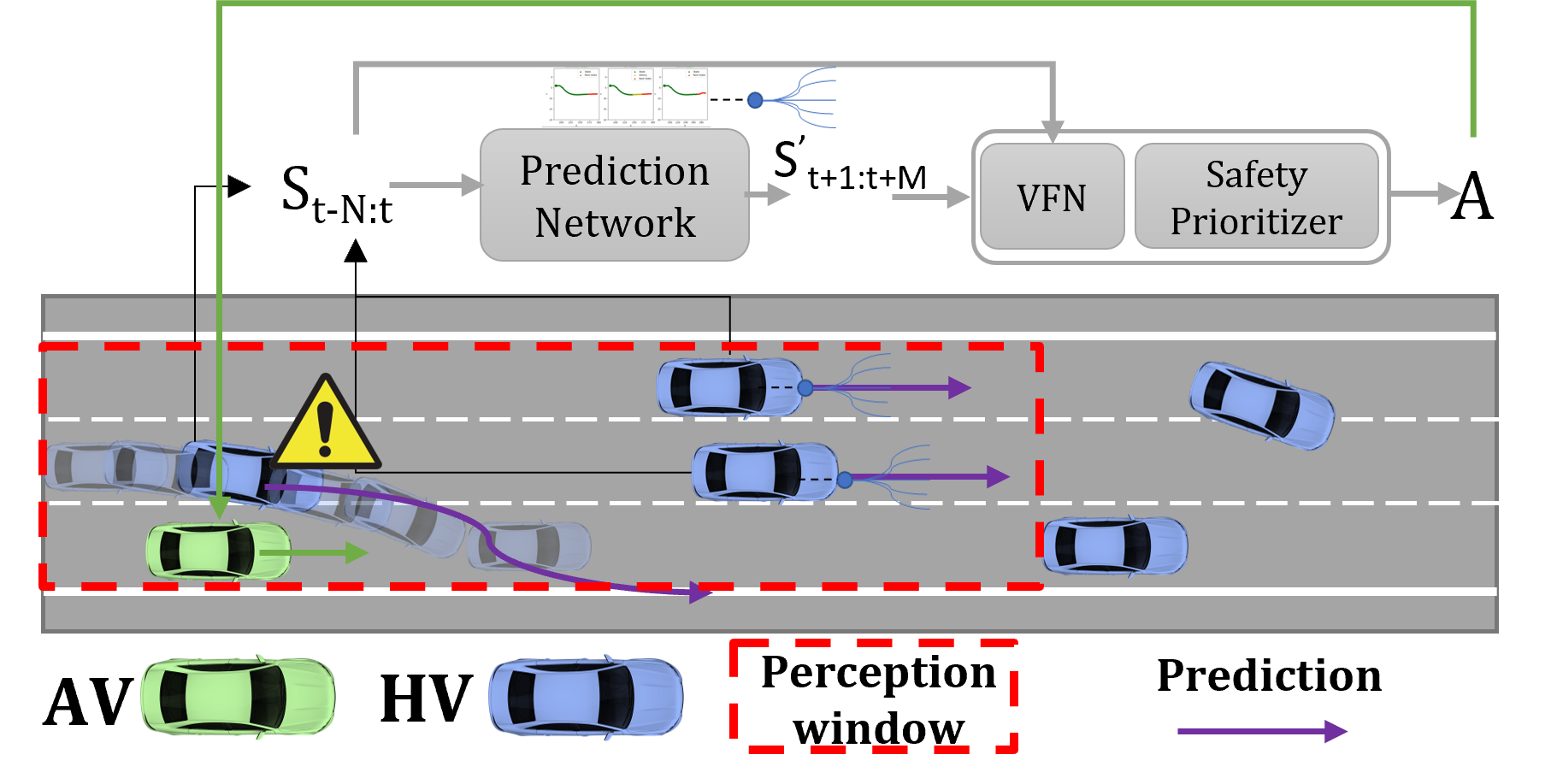}
  \caption{\small{It is crucial for AV being capable of anticipating future situations of other vehicles using spatial and temporal information. Decision-making by AVs can be enhanced by anticipating the intentions of other agents, which is crucial in complex scenarios and safety-critical situations. The figure depicts an AV (in green) and HVs (in blue) with corresponding predictions within the AVs perception window.}}
\label{fig:mainintutition}
\end{figure}

CAVs use Vehicle to Vehicle (V2V) communication to acquire precise situational awareness~\cite{aoki2020cooperative, toghi2019spatio, shah2019real, shah2020rve, shah2022enabling}. We highlight the advantage of CAVs to improve AVs' robustness and safety in two main directions, first, to overcome the limitations of local sensors, and second, to allow coordination among AVs. An effective and reliable means of communication among agents can facilitate AV-HV coordination. AVs and HVs equipped with such reliable vehicular communication can coordinate, improving safety and efficiency~\cite{aoki2020cooperative}. However, even in the presence of perfect communication, HV-AV interactions are still challenging as the behavior and intentions of HVs are still unknown, despite the vehicles' ability to perceive or share information in the communication network. 
Anticipating agents’ behaviors and actions is an important part of real AVs, particularly in mixed autonomy environments. Due to the significance, prediction and HV behavior modeling is an active area of research~\cite{salzmann2020trajectron++, mozaffari2020deep, ivanovic2019trajectron, ivanovic2022injecting, xie2020learning}.

\begin{figure*}[t]
  \centering
  \includegraphics[width=.99\textwidth]{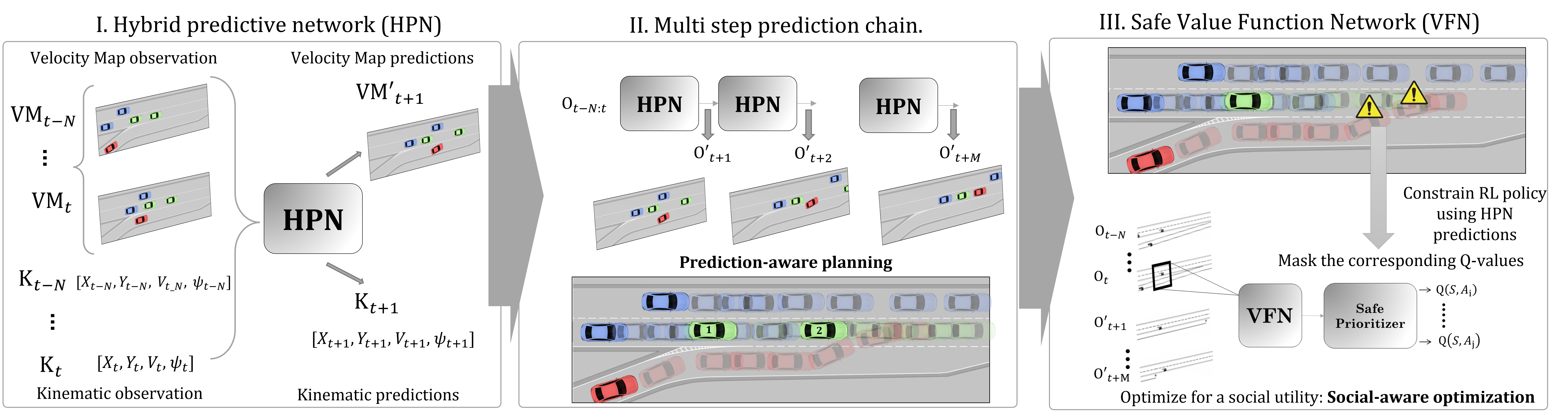}
  \caption{\small{An overview of our prediction-aware and social-aware cooperative driving approach for Multi-agent Cooperative Reinforcement Learning to improve the safety of CAVs. The proposed decentralized RL architecture employs an HPN, which gives AVs the ability to anticipate future states, which is used by the VFN to optimize for social utility and by the safety prioritizer to avoid high collision potential actions.}}
  \label{fig:cover}
\end{figure*}

However, while extensive research has been done in forecasting vehicles trajectories for classical AV stack~\cite{salzmann2020trajectron++, mozaffari2020deep, ivanovic2019trajectron, ivanovic2022injecting}; using prediction in the decision-making process has received less attention. Particularly in the domain of cooperative driving, social interactions, and multi-agent reinforcement learning (MARL), it is important to provide AVs with such capabilities, as presented in Figure~\ref{fig:mainintutition},
in which AV decision-making may benefit from anticipating the future states of other vehicles.  Existing literature proposes probabilistic HV modeling~\cite{mahjoub2019representing} learned from human driving data or rule-based or hand-engineering methods to guide the AVs~\cite{rios2016survey}.
These systems frequently have difficulty communicating or negotiating with other vehicles in complex scenarios.
Other approaches use RL~\cite{lin2020anti, toghi2021altruistic,toghi2021cooperative} and consider social interactions~\cite{toghi2021altruistic,toghi2021cooperative} of AVs-HVs, proposing AVs that can learn from experience and influence HVs, while optimizing a social utility function that benefits all vehicles on the road. However, these approaches do not consider the evolution of the environment into the future and lack the capability of anticipating future states that can be used for decision-making.

Towards this end, we study how CAVs can leverage prediction and social awareness in RL decision-making, to improve safety and efficiency. Therefore, we propose the integration of those important components for AVs, i.e., prediction and social navigation.
Figure~\ref{fig:cover} presents an overview of our approach. First, we propose a Hybrid Predictive Network (HPN) that intends to provide AVs the ability to predict other agents’ potential future, as illustrated in Figure~\ref{fig:cover}-I. 
Second, we use the HPN in a multi-step prediction chain that delivers a window of predicted observations to the value function network (VFN) (as illustrated in Figure~\ref{fig:cover}-II). Finally, the safe VFN relies on a decentralized cooperative RL architecture that optimizes for a social utility and uses expressive velocity map predictions as part of the input states and interpretable kinematic predictions for a safety prioritizer. The safety prioritizer uses the kinematic predictions from the multi-step HPN to constrain the RL policy to ensure safety decision-making by masking the Q-states that produce unsafe results, as shown in Figure~\ref{fig:cover}-III.
We evaluate our prediction-aware and social-aware AV with related approaches under a variety of settings and show that, given the ability to forecast future observations, the AVs can use our proposed approach to improve safety, effectiveness, and overall traffic flow.

Our contributions are as follows:
\begin{itemize}
\item We present a prediction-aware, social-aware decentralized cooperative RL framework and formalize the altruistic cooperative driving problem as a partially observable stochastic game (POSG).
\item We present a hybrid predictive network that provides AVs the ability to anticipate future observations and use it in a multi-step prediction chain that delivers multiple future observations to the value function network. 
\item We describe a robust safety prioritizer that uses interpretable kinematic predictions from the HPN to minimize future high-risk actions, constraining the RL policy to ensure safe decision-making, enhancing awareness of the imminent hazards, reducing collisions, and accelerating the learning process.
\end{itemize}

\section{Related Work}
\label{sec:relatedworks}
\subsection{Reinforcement Learning and Social Navigation}
\noindent Current research in social navigation has demonstrated the importance of AVs as social actors and the benefits of AV-HV coordination~\cite{pokle2019deep}. A method for modeling and forecasting human behavior in situations involving multi-human interactions in highly multi-modal situations is proposed in~\cite{ivanovicgenerative}. HV models are learned from demonstration using inverse RL in~\cite{kuderer2015learning} and~\cite{sadigh2018planning}. Similarly, a centralized stochastic game model approach is presented in~\cite{hadfield2016cooperative}. The authors in~\cite{trautman2010unfreezing} and~\cite{nikolaidis2015efficient} proposed a shared reward function to enable cooperative trajectory planning for robots and humans. Sadigh~\etal presents a strategy based on imitation learning, allowing  AVs to influence HVs~\cite{sadigh2016planning}.

At the traffic level, the importance of coordination and the benefit of using AVs to guide traffic have also been investigated. Wu~\etal~\cite{wu2018stabilizing} examines AVs' ability to stabilize a system of HVs and presents conditions under which enforcing safety constraints on the AVs while stabilizing the traffic improves the overall traffic performance. Similar works have highlighted the potential of influencing HVs and how AVs can guide the traffic flow~\cite{wu2018stabilizing,lazar2019learning}.

Recent works focus on optimizing traffic networks in mixed autonomy to reduce traffic congestion and improve safety and a model of vehicle flow is presented in~\cite{biyik2021incentivizing} in which the planner optimizes for a social goal while improving traffic efficiency. The vehicle routing problem is studied in~\cite{li2021learning}, which proposes an innovative learning-augmented local search system using a transformer architecture to mitigate the problem. In contrast to previous works, we do not rely on human cooperation, and secondly, our AVs incorporate prediction and planning to improve decision-making in cooperative driving.

\subsection{Safety in Autonomous Vehicles} 

\noindent Safety is a priority for real-world adoption of AVs~\cite{Learning_safe,Safety_ECC}. As a result of the complexity of the driving task, safety concerns have been raised, such as: How can we prioritize AV safety in the face of uncertainty? and How can we train RL agents that prioritize AV safety?

Although priority should be given to safety, it often comes at the expense of effectiveness. Consider the following scenario: you decide to pass a car that is moving slowly in front of you. Overtaking a slower car poses a risk because the driver may abruptly change lanes and cause an accident. The only way to ensure safety is to avoid overtaking. Following this and similar arguments, it becomes clear that the only condition that completely guarantees safety is avoiding driving. As a result, the goals of efficiency and safety are frequently conflicting. 

AVs based on RL can raise safety concerns as they can select unsafe actions due to function approximation~\cite{li2018safe}. To improve safety, authors in~\cite{wang2019lane} propose a rule-based system for evaluating controller decisions and masking collision-causing actions. Cameron~\etal investigate how humans can supervise agents to achieve an acceptable level of safety~\cite{hickert2021cooperation}.
Others use a pure reward-shaping strategy, however, in this case, safety is not prioritized and the AVs are susceptible to select dangerous actions~\cite{li2018safe,nageshrao2019autonomous}. 
To overcome this problem, authors in~\cite{li2018safe} present the concept of safe RL, which aims to increase safety in unobserved environment conditions. 
A short-horizon safety supervisor is proposed in~\cite{nageshrao2019autonomous} to replace unsafe actions with safer ones. A Q-masking strategy is presented in~\cite{mohammadhasani2021reinforcement} to prevent collisions by deleting actions that might lead to a crash. Authors in~\cite{chen2021deep} propose a safety supervisor that considerably decreases crashes~\cite{chen2021deep}.

To improve safety we employ a safety prioritizer that uses kinematic predictions from the multi-step prediction network to look ahead and avoid imminent collisions by masking high-risk actions in the short-horizon.

\subsection{Behavior Modeling and Prediction}
\noindent Human behavior is difficult to predict, and human decision-making is governed by inherently unobservable cognitive processes. 
The current works on driver behavior and social navigation approaches agents' coordination by either modeling driver behavior~\cite{brown2020taxonomy,ivanovicgenerative,mahjoub2019representing,Augmented} or simplifying and making assumptions about the nature of agent interactions~\cite{lauer2000algorithm,omidshafiei2017deep}. 
Other works on driver behavior modeling consider data mining~\cite{constantinescu2010driving}, driver attributes~\cite{beck2014distress}, graph theory~\cite{chandra2020cmetric}, or game theory~\cite{schwarting2019social}. 

In vehicular safety applications, the use of kinematic equations by CAVs for the prediction of neighboring vehicles' positions and trajectories in short time horizons is a common approach. Dynamic and kinematics-based solutions are used in~\cite{helbing1995social}. These methods usually consider the vehicles to be rigid point masses and assume that the longitudinal velocity, acceleration, or other motion moments are constant, which are frequently accepted assumptions for prediction by conventional vehicle manufacturers. 
Amongst these kinematic models, the constant speed (CS) or acceleration (CA) model has more popularity for the prediction of road participants' position and speed in the cooperative vehicle safety application domain~\cite{Parker2007,Baek2017,Painter1990,Chu2010,hnmahjoub:syscon19}. 

Numerous methods have defined trajectory prediction as a regression problem, and potent methods like Inverse Reinforcement Learning~\cite{lee2016predicting}, Recurrent Neural Networks~\cite{morton2016analysis, alahi2017learning} and Gaussian Process Regression and Gaussian Mixture Models ~\cite{9774935,9644629,das2010block, wang2007gaussian}, have been successfully applied in different applications. Authors in \cite{guan2019intelligent} present a LSTM model to forecast vehicle's trajectories.
Other works leverage non-parametric Bayesian approaches to predict fundamental patterns of observable time series. Particularly, within the non-parametric bayesian inference approaches, Gaussian Process (GP) has demonstrated a significant performance~\cite{hnmahjoub:syscon19}. Because of recent advances in deep generative models, generative approaches have been widely used~\cite{goodfellow2014generative, sohn2015learning}. The majority of works in this domain use Autoencoders, such as Conditional Variational Autoencoder, Recurrent Variational Autoencoder, or Generative Adversarial Network ~\cite{goodfellow2014generative, sohn2015learning}.

Differently, we propose a combined approach to predict velocity map images directly using a predictive autoencoder architecture and interpretable kinematics predictions using a GP solution. We use these predictions to improve decision-making, thereby integrating social navigation with prediction, which are crucial elements for AV navigation.


%
\section{Preliminaries and Problem Formulation}
\label{sec:preliminaries}

\subsection{Autoencoder}
An Autoencoder (AE) is a neural network that is trained in an unsupervised approach to minimize the reconstruction error. The AE learns important features that allow reconstructing the original input and its architecture is generally divided into two main components: an encoder and a decoder. Following a similar approach, an AE can be trained for a prediction task assuming that we have the state at time $t$ and the corresponding state at time $t+1$. 
Formally, the encoder maps the input $\boldsymbol{x}$ to a latent feature representation $\boldsymbol{z}$ denoted by $z = f_{w_e}(x)$. The decoder uses the latent representation $\boldsymbol{z}$ to obtain a reconstruction $\boldsymbol{y}$ of the input $\boldsymbol{x}$ , denoted by $y = f_{w_d}(z)$. The reconstruction error, i.e., the difference between $\boldsymbol{x}$ and the reconstruction $\boldsymbol{y}$ is used as the objective function.

\subsection{Gaussian process}
Gaussian processes (GP) are frequently used to predict future trajectory time series by regressing the observed time-series realizations from history, capturing the distinct patterns as they emerge in the data, making GP a useful tool for detecting patterns in time series~\cite{Rasmussen:GP, mahjoub2019representing}. 
When using GP to forecast future trajectories, the set of $m$ observed values is represented by an m-dimensional multivariate Gaussian random vector, described by an $m \times m$ covariance matrix and a $m$ mean vector. This covariance matrix, often known as the GP kernel, is the foundation upon which GP detects and anticipates the underlying behavior of time series based on their recorded history. The fundamental GP components can be expressed mathematically as follows:

\begin{subequations} \label{bounds}
    \begin{gather}
        \label{egu:f}
        f(t) \sim gp \big(m(t), k(t,t')\big),\\ 
        \label{egu:X}
        \{X_{i}\}_{i=1,2,...,m} = \{f(t_i)\}_{i=1,2,...,m} \sim \mathcal{N}(\mu,\,\Sigma),\\
        \label{egu:mu}
        \mu = \big[m(t_1), ..., m(t_m)\big]^T,\\
        \Sigma_{ij} = \kappa(t_i, t_j) \ \forall i,j \in \{1,2, ..., m\}
    \end{gather}
\end{subequations}

where $X_{i}$, $f(t)$, $m(.)$, and $\kappa(.,.)$ are the samples of the vehicles' state, observed or to be predicted at the time $t_i$, the unknown underlying function that the vehicles' states are sampled from, the mean and the covariance functions, respectively.

In this work, we leverage GP to improve kinematics prediction, and instead of working directly with the position time series, our GP inference algorithm treats the vehicles' heading and longitudinal speed as two independent time series that are regressed using GPs, and then using the predicted heading and longitudinal speed, the vehicles' positions are calculated.
A model built using a non-parametric Bayesian inference framework dynamically adapts its complexity to the observed data, preventing overly complicated models yet catching unexpected patterns in the data as they emerge.

\subsection{Partially Observable Stochastic Games (POSG)}
\noindent In this section we present the notation for our RL based altruistic cooperative driving problem, defined as a POSG denoted by $\langle \mathcal{I}, \mathcal{S}, P, \{ \mathcal{A}_i \}_{i\in\{1,...,N\}}, \{ \mathcal{O}_i \}_{i\in\{1,...,N\}}, \{ R_i \}_{i\in\{1,...,N\}} \rangle ,\gamma $.
At a given time $t$ each agent $i \in \mathcal{I}$ perceives the environment and receives a partial observation $o_i: \mathcal{S} \rightarrow \mathcal{O}_i$, considering the observation $o_i$ and its policy $\pi_i: \mathcal{O}_i \times \mathcal{A}_i \rightarrow [0, 1]$, the agent takes an action $a_i \in \mathcal{A}_i$ and transits to the state $s' \in \mathcal{S} $ based on the transition probability $P(s'|s, a)$ and receives a local reward $r_i\in \mathcal{R}$. Each agent $i$ seeks to find an optimal policy $\pi^*: \mathcal{S} \rightarrow \mathcal{A}$, that maximizes the sum of future rewards $r_i\in \mathcal{R}$,
i.e., $\pi^*(s) = \arg\max_a Q^* (s,a) $, where, $Q^\pi(s,a) \coloneqq \mathbb{E}_{\pi} [\sum_{k=0}^\infty \gamma^k R_k(s,a) |s_0=s, a_0=a]$, in which, $\gamma \in [0,1)$ is the discount factor. The optimal action-value function can then be obtained using the Bellman optimality equation, $Q^*(s,a) = \mathbb{E} \left[ R(s,a) + \gamma \max_{a'} Q^*(s',a') |s_0=s, a_0=a \right]$.

\subsection{Deep Q-Network}
\noindent Deep Q-network (DQN) and Double Deep Q-Network (DDQN) have been widely used in RL problems. DDQN regularly samples data from a buffer in order to calculate an estimate of the Bellman error, which is denoted by the following formula:

\begin{equation}
\label{equ:loss2}
\mathcal{L}(\textbf{w}) = \mathbb{E}_{s,a,r,s' \sim \mathcal{RM}}[( Target - \Tilde{Q}(s,a;\textbf{w}))^2]
\end{equation}
\vspace{-5pt}
\begin{equation}
\label{equ:DDQNtarget}
Target = R(s,a) + \gamma \Tilde{Q}(s',\underset{a'}{\arg\max} \Tilde{Q}(s',a';\textbf{w});\hat{\textbf{w}}))
\end{equation}

Following this, the DDQN algorithm learns an approximate action-value function ($\Tilde{Q}(.)$) by performing gradient descent steps as $\textbf{w}_{i+1} = \textbf{w}_i - \alpha_i \hat{\nabla}_\textbf{w} \mathcal{L}(\textbf{w})$, on the loss $\mathcal{L}$. Here, $\textbf{w}$ represents the online network weights and $\hat{\textbf{w}}$ represents the target network weights (updated at frequency $Target_{update}$). The experience replay buffer ($RM$) is used to generate training samples $(s, a, r, s')$, which are randomly drawn to protect from correlated observations and non-stationary data distribution.

\subsection{Driving Scenarios and Behaviors}
\noindent We study the performance of our framework on multiple HV behaviors and scenarios. We design a set of scenarios, $\mathcal{F}$ such as straight highway, highway exiting, highway merging, intersection, and roundabout scenarios, defined as $f_h, f_e, f_m, f_i, f_r \in \mathcal{F}$ correspondingly. 
Using these scenarios, we train AVs that are social-aware by using an altruistic reward that embedded Social Value Orientation (SVO) in the AVs. Properly, we describe social preferences (altruism or egoism) by the AV's SVO angular phase $\phi$~\cite{toghi2021altruistic,toghi2021social}. To simulate diverse behaviors we compute the appropriate parameter values that simulate the desired behaviors. We compute the HV driver parameters ($\mathcal{P}$) and based on the parameters ($\mathcal{P}$) generate a set of behaviors $\mathcal{B}$, i.e., conservative, moderate and aggressive, $b_c,b_m,b_a \in \mathcal{B}$ used within the simulator~\cite{chandra2020cmetric,valiente2022robustness}.
A mixed behavior scenario is obtained by sampling from the behaviors in $\mathcal{B}$.

\subsection{Problem Formulation}
\label{sec:problem_formulation}

\noindent In this work, we focus on prediction-aware planning for altruistic cooperative driving. We assume our scenarios contain a set of AVs $i_i \in \mathcal{I}$ and HVs $h_k \in \mathcal{H}$, with diverse SVO.  We assume that AVs are connected and perceive a partial observation of the environment $\Tilde{\textbf{o}}_{i} \in \widetilde{\mathcal{O}}_i$, perceiving a subset of vehicles ${\mathcal{C}} = \widetilde{\mathcal{H}} \cup \widetilde{\mathcal{I}}$, i.e., a subset of HVs $\widetilde{\mathcal{H}} \subset \mathcal{H}$ and AVs $\widetilde{\mathcal{I}} \subset \mathcal{I}$. We study the following problem: How AVs can leverage prediction in decision-making to learn optimal cooperative policies $\pi^*(s)$ in a mixed-autonomy environment under different HVs behaviors $b \in \mathcal{B}$ and scenarios $f \in \mathcal{F}$. 

The RL-based altruistic cooperative driving problem is formalized as a POSG as described previously, attempting to obtain optimal policies that produce socially advantageous outcomes. To formalize our prediction problem, let us represent our state at time $t$ for the vehicle (car) $c$, $c \in {\mathcal{C}}$, as $s^c_t$ and let $s_t = s^{1,...,|C|}_t$ represent the state for all the vehicles within the perception range. We assume that our state $s_t$ consists of a stack of $N$ past observations and $M$ future hypotheses, accounting for temporal and prediction information, i.e., $s_t = [\Tilde{\textbf{o}}_{t-N:t} ,\Tilde{\textbf{o}}_{t+1:t+M}^{\prime}]$ for all the vehicles within the local observation. 
The prediction system takes as input the previous observations $\Tilde{\textbf{o}}_{t-N:t}$ and aims to produce $\Tilde{\textbf{o}}_{t+1:t+M}^{\prime}$. We note that this is the general notation, and in our framework, $s_t$ is not just the vehicle trajectory, but a combination of vehicle kinematic trajectory and a velocity map, the  details of which are presented in the following sections. 

The previous ($\Tilde{\textbf{o}}_{t-N:t}$) and anticipated observations ($\Tilde{\textbf{o}}_{t+1:t+M}^{\prime}$) are used to learn an optimal policy at a given state $s_t$, $\pi^*: \mathcal{S} \rightarrow \mathcal{A}$
The goal of this work is to train prediction-aware and social-aware AVs that can drive safely in a mixed-autonomy scenario.

\vspace{-0.2cm}
\section{Prediction-aware altruistic cooperative driving}
\label{sec:solution}
\noindent The POSG becomes significantly more complex in the presence of HVs since their behavior is difficult to predict and change over time. Therefore, predicting HV behavior is crucial for AVs' in a mixed-autonomy environment. On the basis of this insight, we develop a framework that combines prediction and planning. We propose a predictive network that provides predictions to the planner, and the planner learns to use those predictions for decision-making. The prediction networks give the AV the capability to anticipate the future, and the VFN embeds the predictions and learns the inter-agent relations while optimizing for a social utility.

Our approach uses the HPN that provides possible future observations. Then the HPN is used in a multi-step prediction chain that produces multiple possible future observations to the VFN. Finally, the VFN is trained to optimize a social utility within the RL framework. The VFN outputs Q-values, that are masked by a safety prioritizer, constraining the RL policy to a safe action space.
The outline of our framework is presented in Figure~\ref{fig:cover}. 

The two main sub-systems are the HPN and the VFN, where HPN is a predictive autoencoder and VFN is a 3D convolutional neural network (CNN). 
We hypothesize that the combination of prediction (HPN) and decision-making (VFN) improves the AVs' ability to learn to navigate complex scenarios. The input of the system is the hybrid spatio-temporal state representation, i.e., VelocityMaps and kinematic state and the output are the action-values, and after the unsafe actions are masked, the action with the highest Q-values is selected ($a = \max_{a' \in \widetilde{\mathcal{A}}_{safe} } Q(s,a';\textbf{w})$ at the given state $s \in \mathcal{S}$). To encourage the required safe social behavior in the AVs, we design a suitable reward function.

\subsection{Action and State Space}
\label{sec:spaces}

\textbf{Action Space:}
This study aims to investigate interactions between agents and between AVs and HVs.
As a result, we decide to choose the action-space as a collection of discrete meta-actions $a_i \in \mathcal{A}_i$ at an abstract level, and the abstract actions are transformed into control signals. We specifically choose a set of actions ($a_i$) as follows:
\begin{equation}
\label{equ:action_space}
    a_i \in \mathcal{A}_i =
    \begin{bmatrix}
        \texttt{Lane Left}\\
        \texttt{Idle}\\
        \texttt{Lane Right}\\
        \texttt{Accelerate}\\
        \texttt{Decelerate}
    \end{bmatrix}
\end{equation}

\textbf{State Space:}
The AVs at every time step $t$ receive a local observation of the environment $\Tilde{\textbf{o}}_{t} \in \widetilde{\mathcal{O}}_t$. As temporal information is crucial for the driving task we incorporate $N$ consecutive observations.
We use VelocityMaps ($VM$) and Kinematic ($K$) information, at time step $t$, each combination of $VM$ and $K$ is an observation from the environment as,
\begin{equation}
\label{equ:obs_space}
    \Tilde{\textbf{o}}_{t} \in \widetilde{\mathcal{O}}_t =
    \begin{bmatrix}
        \texttt{$VM_t$}\\
        \texttt{$K_t= X_{t},Y_{t},V_{t},\Psi_{t}$} 
    \end{bmatrix}
\end{equation}

The kinematic information is included to explicitly incorporate the movement data, which helps the training process, and also serves to obtain accurate Kinematic prediction for the safety prioritizer. Additionally, as anticipating futures states is also important for decision-making in complex scenarios, the prediction chain generates a sequence of $M$ hypotheses from the observations that provide information on how the environment could probably evolve into the future. We combine $N$ consecutive past observations and $M$ hypotheses from the prediction network to construct a more useful state. Therefore, our state consists of a stack of $N$ past observations and $M$ futures hypothesis, accounting for temporal and prediction information, i.e., $s_t = [\Tilde{\textbf{o}}_{t-N:t} ,\Tilde{\textbf{o}}_{t+1:t+M}^{\prime}]$.

The $VM$ information incorporates the relative vehicle’s speed in pixel values~\cite{toghi2021social}.
The $K_t$ is a matrix in which the rows are the number of vehicles included in the observation and the columns contain the kinematics information for each vehicle.
The kinematics information for each vehicle $c$, i.e., $k_t^c$, is a 4-dimensional vector encoding the vehicle’s position, velocity and heading, and it is formed by the kinematics of the vehicle $x,y,v,\psi$, where $(x,y)$ represents the vehicle position, $v$ is the longitudinal speed, $\phi$ is the heading, and $(\dot{x} , \dot{y})$ are computed as $\dot{x} = v\cos{\phi}, \dot{y} = v\sin{\phi}$.
The $K_t$ embeds the kinematics of surrounding vehicles, and it
includes the kinematics information $k_t^c$ for all the $c \in {\mathcal{C}}$ vehicles, in addition to the ego vehicle, i.e., $K_t = \big[ k_t^{ego}, k_t^{1}, k_t^{2},..., k_t^{|C|} \big]^\top$. Each row $r$ of $K_t$ matrix contains the kinematics information for the vehicle $c$, $k_t^c = [x_t^c,y_t^c,v_t^c,\psi_t^c]$. 

\subsection{Reward Function}
\label{sec:reward}
\noindent Inspired by the work in sympathy and cooperation for autonomous driving~\cite{toghi2021altruistic}\cite{toghi2021social}, we design a reward function that takes into account the traffic metrics, social and individual rewards of AVs and HVs. Therefore, we separate the reward function for each agent $I_i \in \mathcal{I}$ in two terms: an egoistic reward $R^{\mathrm{ego}}$, and a social reward $R^{\mathrm{social}}$, as

\begin{subequations} \label{decentralizedreward}
    \begin{gather}
        \label{egu:r1}
        R_i(s, a) =R^{\mathrm{ego}}+R^{\mathrm{social}},\\ 
        \label{egu:r2}
        R^{\mathrm{ego}} = \cos \phi_i r_i(s, a),\\
        \label{egu:r3}
        R^{\mathrm{social}} = \sin \phi_i \sum_j r^{\mathrm{AV}}_{i, j} (s, a) + \sin \phi_i \sum_k r^{\mathrm{HV}}_{i, k} (s, a)
    \end{gather}
\end{subequations}

in which $i \in \mathcal{I} $, $j \in (\widetilde{\mathcal{I}} \setminus \{I_i\})$, $k \in \widetilde{\mathcal{H}}$. The ego vehicle's reward is defined by $r_i$ and the angle $\phi$ allows the adjustment of the level of the egoistic and social components.

The $R^{\mathrm{social}}$ term considers the social utility of the $k$ HVs and $j$ AVs for the agent $i$, i.e., $r^{\mathrm{HV}}_{i, k}$ and $r^{\mathrm{AV}}_{i, j}$ defined as $r^{\mathrm{HV}}_{i, k} = \frac{1}{d_{i,k}^\lambda} \sum_m \omega_m x_m$ and $r^{\mathrm{AV}}_{i, j} = \frac{1}{d_{i,j}^\lambda} \sum_m \omega_m x_m$, respectively, where $m$ is the set of traffic metrics that have been taken into account in the utility of the vehicle (crashes, speed, and distance traveled), $x_m$ represents the $m$ metric, and $w_m$ represents the weights (metrics importance). The terms $d_{i,k}/d_{i,j}$ represent the gap between the ego-AV and the associated HV/AV and $\lambda$ is a hyperparameter that sets the importance of neighboring vehicles.

\subsection{Architecture.}
\noindent The proposed prediction-aware planning architecture is presented in Figure~\ref{fig:cover} and consists of the Hybrid predictive network (HPN, Figure~\ref{fig:cover}-I), the multi-step prediction chain (Figure~\ref{fig:cover}-II), the value function network (VFN, Figure~\ref{fig:cover}-III), and the safety prioritizer. 
The HPN (as shown in Figure~\ref{fig:hpn}) serves as a predictive autoencoder network. It takes as input the history of observations at time $t$, i.e., $\Tilde{\textbf{o}}_{t-N:t}$ and produces a predicted observation at time $t+1$, i.e., $\Tilde{\textbf{o}}_{t+1}^{\prime}$. 
The prediction chain is a multi-step prediction chain that uses the HPN in a chain to produce a set of $M$ hypotheses. It takes a history of observations at time $t$, i.e., $\Tilde{\textbf{o}}_{t-N:t}$ and produces a set of $M$ predicted observations, i.e., $\Tilde{\textbf{o}}_{t+1:t+M}^{\prime}$ for the VFN. 
Prediction-aware planning is made possible by combining prediction (HPN) and decision-making (VFN), which improve driving performance in challenging situations. The details of the architecture are presented in the following sections.

\subsubsection{Hybrid predictive network (HPN)}
\begin{figure}[t!]
  \centering
  \includegraphics[width=.48\textwidth]{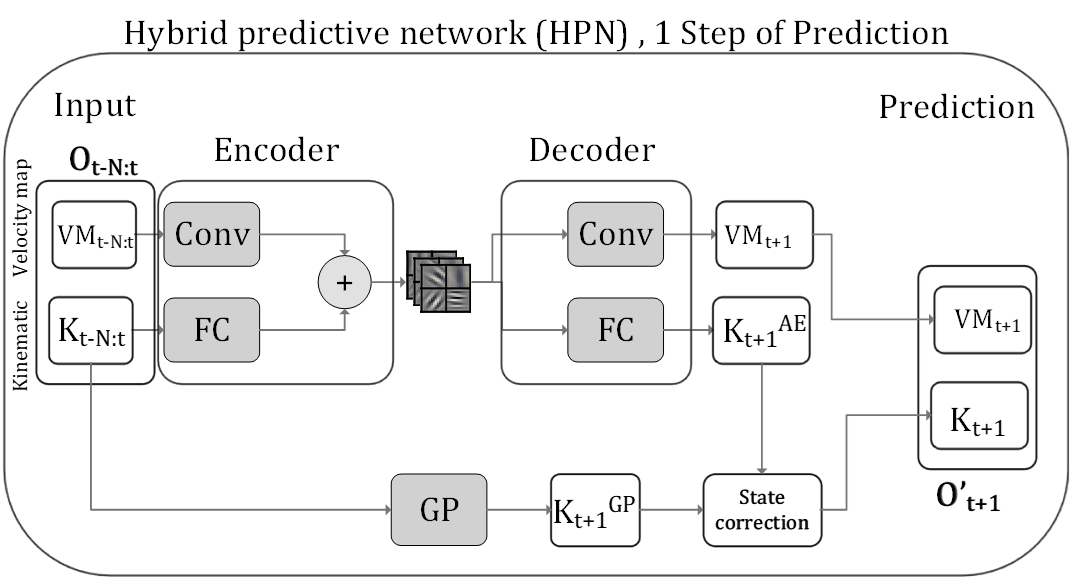}
  \caption{\small{Architecture of the hybrid predictive network (HPN) for one prediction step. Using HPN in a chain will give AVs the ability to predict future observations.}}
\label{fig:hpn}
\end{figure}

The HPN (Figure~\ref{fig:hpn}) is a prediction autoencoder network, it uses the sequence of $N$ observations at time $t$, i.e., $\Tilde{\textbf{o}}_{t-N:t}$ and outputs a predicted observation at time $t+1$, i.e., $\Tilde{\textbf{o}}_{t+1}^{\prime}$. The HPN consists of a symmetric encoder-decoder architecture. The encoder consists of 3 convolutional layers with 3x3 filters, with 32, 64, and 64 feature maps. The encoder takes as input the history of observations ($\Tilde{\textbf{o}}_{t-N:t} $), where each observation consists of a velocity map image ($VM_t$) and the Kinematic matrix ($K_t$). The $VM$ from $t-N:t$ are passed through the 3-convolutional layers and the $K$ vectors from $t-N:t$ are passed through 2-FC (fully connected) layers with 128 hidden units, whose final layer contains the same number of hidden units as in the convolution network (CNN) output. The outputs ($VM$ features and $K$ features representations) are combined using element-wise addition operation. The decoder consists of a symmetric version of the encoder, i.e., a deconvolutional network with 3-convolutional layers and 2-FC layers. The convolutional layers produce a prediction for the next $VM_{t+1}^{\prime}$ and the FC layers produce the prediction for the next $K_{t+1}^{\prime}$. The CNN encoder is designed to extract important spatial information of the input $VM$ image. 
The predictive autoencoder is trained by minimizing the Mean Squared Error (MSE) between the prediction $\Tilde{\textbf{o}}_{t+1}^{\prime}$ and the target $\Tilde{\textbf{o}}_{t+1}$.

Although the AE provides kinematic predictions, we found that an indirect hybrid GP prediction approach to correct the kinematic predictions provides better results. Our findings are based on previous works that show how a GP-based prediction system is powerful for accurate kinematic predictions and often performs better than other models like AE and LSTM models~\cite{mahjoub2019representing}\cite{9644629}\cite{Safety_ECC}.

Therefore, while we use the predictive AE to predict the next $VM_{t+1}^{\prime}$ image and $K_{t+1}^{\prime}$ state, we correct the kinematic state $K_{t+1}^{\prime}$ using a GP approach to improve kinematics prediction. We particularly find that accurate kinematics prediction are important for the safety prioritizer that uses the predictions to constrain the RL policies to a safer space. In this work, instead of directly using the position time series ($x_{t-N:t},y_{t-N:t}$) for each vehicle, our GP inference algorithm treats the vehicles' heading ($\psi_{t-N:t}$) and speed ($v_{t-N:t}$) as two independent time series that are regressed using GPs, and then calculates the vehicles' position ($x_{t+1:t+M}^{\prime},y_{t+1:t+M}^{\prime}$) using the predicted heading and speed. We call this approach GP-indirect prediction and present more details in the following sections.

For each vehicle $c$, the GP prediction algorithm takes as input the history of the 4-dimensional kinematic vector ($x_{t-N:t},y_{t-N:t},v_{t-N:t},\psi_{t-N:t}$, ), uses the heading ($\psi_{t-N:t}$) and speed ($v_{t-N:t}$) time series to predict their future values ($\psi_{t+1:t+M}^{\prime},v_{t+1:t+M}^{\prime}$). Then after modelling speed and heading, the future position of the vehicle, i.e., $x_{t+1:t+M}^{\prime},y_{t+1:t+M}^{\prime}$ is computed as follows, 

\begin{subequations} \label{gp_v}
    \begin{gather}
        \label{egu:v1}
       \{V_i\}_{i=1,2,...,m} = \{f_{speed}(t_i)\}_{i=1,2,...,m} \sim \mathcal{N}(\overline{\mu_v},\,\Sigma_v),\\ 
        \label{egu:v2}
       \overline{\mu_v} = m_v(t_i); \ \Sigma_{v_{i,j}} = k_v(t_i, t_j) \ \forall i,j \in \{1,2, ..., m\},\\
        \label{egu:v3}
         f_{speed}(t) \sim gp (m_v(t), k_v(t,t'))
    \end{gather}
\end{subequations}

\begin{subequations} \label{gp_h}
    \begin{gather}
        \label{egu:h1}
       \{\psi_i\}_{i=1,2,...,m} = \{f_{heading}(t_i)\}_{i=1,2,...,m} \sim \mathcal{N}(\overline{\mu_\psi},\,\Sigma_\psi),\\ 
        \label{egu:h2}
       \overline{\mu_\psi} = m_\psi(t_i); \ \Sigma_{\psi_{i,j}} = k_\psi(t_i, t_j) \ \forall i,j \in \{1,2, ..., m\},\\
        \label{egu:h3}
         f_{heading}(t) \sim gp (m_\psi(t), k_\psi(t,t'))
    \end{gather}
\end{subequations}

\begin{subequations} \label{xy}
    \begin{gather}
        \label{egu:x}
         x_{t+1}^{\prime} = x_t  + \int_{t}^{t+1} f_{speed}(t)\, cos(f_{heading}(t))\, dt ,\\ 
        \label{egu:y}
        y_{t+1}^{\prime} = y_t  + \int_{t}^{t+1} f_{speed}(t)\, sin(f_{heading}(t))\, dt 
    \end{gather}
\end{subequations}

From the output of the GP model, the 4-dimensional kinematic vector ($ k_{t+1}^{GP} = x_{t+1},y_{t+1},v_{t+1},\psi_{t+1}$) for each vehicle is used to correct the AE kinematic prediction ($k_{t+1}^{AE}$), and the GP prediction is performed for each vehicle in the $K$ matrix (rows of the matrix) and a new matrix is formed with all the predictions at time $t+1$, i.e., $K_{t+1}^{\prime} = \big[ k_{t+1}^{ego}, k_{t+1}^{1}, k_{t+1}^{2},..., k_{t+1}^{|C|}\big]$. The final predicted observation is a combination of the predicted velocity map ($VM_{t+1}^{\prime}$) and the corrected kinematic prediction ($K_{t+1}^{\prime}$) as shown in Figure~\ref{fig:hpn}, i.e., 

\begin{equation}
\label{equ:obs_space_pred}
    \Tilde{\textbf{o}}_{t+1}^{\prime} =
    \begin{bmatrix}
        \texttt{$VM_{t+1}^{\prime}$}\\
        \texttt{$K_{t+1}^{\prime} = X_{t+1}^{\prime},Y_{t+1}^{\prime},V_{t+1}^{\prime},\Psi_{t+1}^{\prime}$} 
    \end{bmatrix}
\end{equation}

\subsubsection{Multi-step prediction chain}
\begin{figure*}[t!]
  \centering
  \includegraphics[width=.98\textwidth]{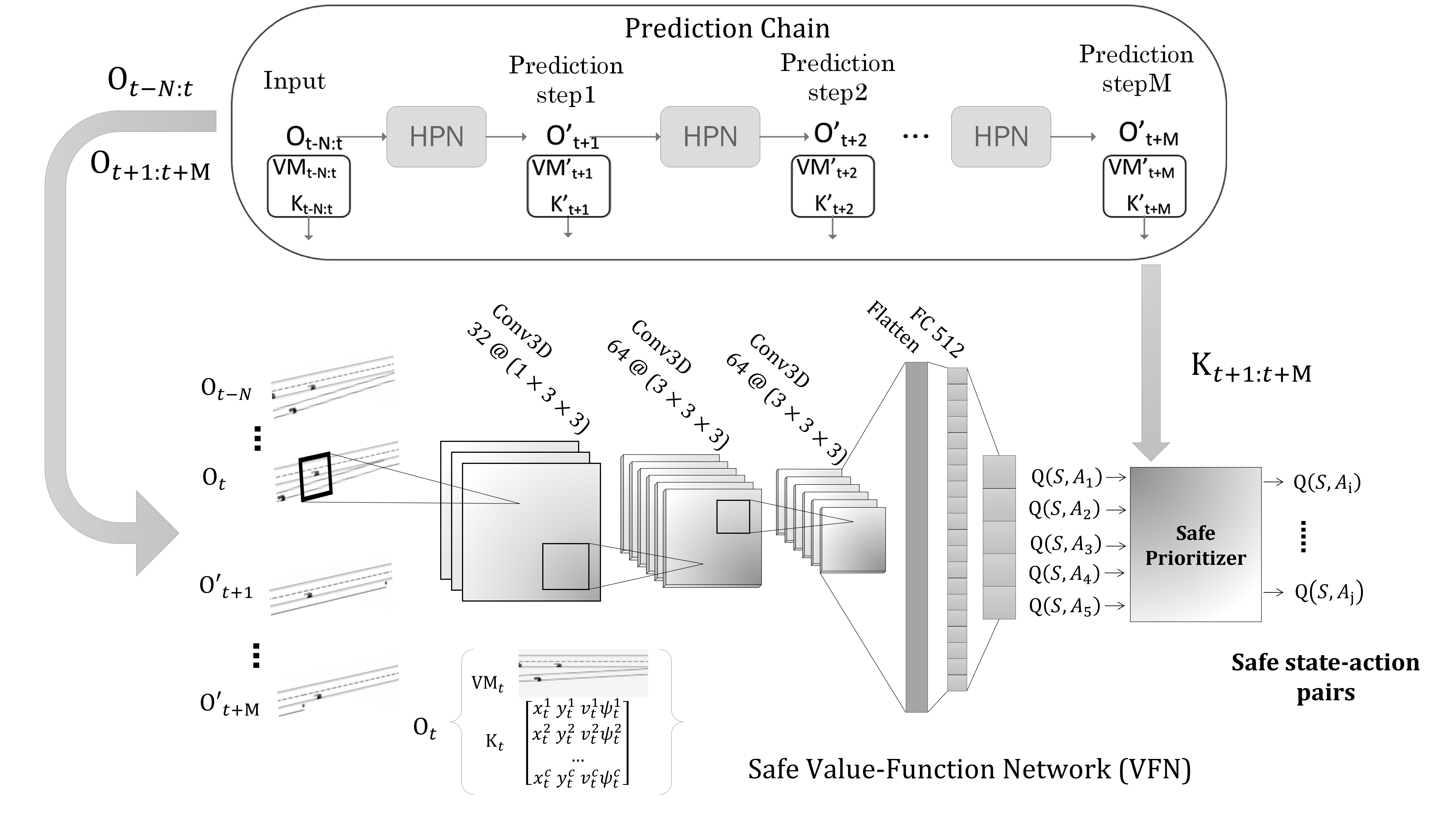}
  \caption{\small{Multi-step prediction chain and Safe Value Function Network (VFN). The prediction chain is a multi-step prediction chain that uses the HPN in a chain to produce a set of $M$ future hypotheses. The VFN is a 3D CNN that acts as a function approximator, it uses temporal information and prediction to improve decision-making.}}
\label{fig:prediction_chain}
\end{figure*}

The prediction chain, as presented in Figure~\ref{fig:prediction_chain} is a multi-step prediction process that uses the HPN (Figure~\ref{fig:hpn}) in a chain to produce a set of $M$ future hypotheses. It takes a history of observation at time $t$, i.e., $\Tilde{\textbf{o}}_{t-N:t}$ and produces a set of $M$ predicted observations, i.e., $\Tilde{\textbf{o}}_{t+1:t+M}^{\prime}$ , as described in algorithm~\ref{alg:hpn}, to compute the input state for the VFN, i.e., $s_t = [\Tilde{\textbf{o}}_{t-N:t} ,\Tilde{\textbf{o}}_{t+1:t+M}^{\prime}]$.
\begin{algorithm}[t]
    \caption{Multi-step prediction chain.} 
    \label{alg:hpn}
    \begin{algorithmic}
    \STATE Input $\Tilde{\textbf{o}}_{t-N:t}$. The sequence of previous observations. 
          \FOR{$t = t$ to $t+M$}
          \STATE Predict $\Tilde{\textbf{o}}_{t+1}^{\prime}$ = HPN ($\Tilde{\textbf{o}}_{t-N:t}$)
          \STATE Save prediction $\Tilde{\textbf{o}}_{t+1}^{\prime}$ and use it for the next step
          \ENDFOR        
    \STATE Output $\Tilde{\textbf{o}}_{t+1:t+M}^{\prime}$. The sequence of predicted observations. 
    \end{algorithmic}
\end{algorithm}

\subsubsection{Safety Prioritizer}
\label{sec:safeprioritizer}
In order to improve safety, we propose a safety prioritizer within our VFN. The safety prioritizer penalizes high-risk actions, thereby reducing imminent crashes. 
If the AVs come into an unexpected situation and based on the output of the VFN, decide to perform a high-risk action, the safety prioritizer will mask the action. The safety prioritizer is comprised of two algorithms, i.e.,~\textbf{Algorithm}~\ref{alg:safecheck} that checks for safe actions and \textbf{Algorithm}~\ref{alg:safeaction} that performs action selection.

\textbf{Algorithm}~\ref{alg:safecheck} verifies if the selected action $a_t$ is safe based on a safety score for $M_{steps}$ of prediction. Algorithm~\ref{alg:safecheck} simulates $I_i$ taking the action $a_t$ and uses the kinematic predictions from HPN, i.e., $\Tilde{\textbf{K}}_{t+1:t+M}^{\prime}$ = HPN ($\Tilde{\textbf{K}}_{t-N:t}$), for all vehicles in the road $ c \in \mathcal{C}$ to compute the time-to-collision ($ttc$) at time $t$, i.e., $ttc_t$ between $I_i$ and all $ c \in (\widetilde{\mathcal{I}} \cup \widetilde{\mathcal{H}}) \setminus \{I_i\}$ using $x,y,v,\psi$, at each prediction step is calculated and the minimum $ttc$ is saved, and using the predicted $ttc$ for all the $M_{steps}$ of prediction ($ttc_{t+1:t+M}$), the $safe_{score}$ is computed. The $safe_{score}$ is a weighted average of the $ttc_{t+1:t+M}$, with exponential decay to give more importance to the short-term predictions. Finally, if $safety_{score}< safe_{th}$ or any of the 
predicted $ttc$ is less than the critical threshold, i.e., $any(ttc_{t+1:t+M})<  critical_{th}$, the action is considered unsafe. The $safe_{th}$ is the safe $ttc$ threshold for possible crash, and $critical_{th}$ is a critical $ttc$ threshold for imminent crash. If the current action is considered unsafe, ~\textbf{Algorithm}~\ref{alg:safeaction} will select another action.

\textbf{Algorithm}~\ref{alg:safeaction} presents the selection of the action. It iteratively verifies the actions using~\textbf{Algorithm}~\ref{alg:safecheck}  and selects a safe action that follows the learned policy. The restricted actions will prevent the agent from engaging in risky behavior during training, resulting in a more balanced learning and efficient sampling.

\begin{algorithm}[t]
    \caption{Action evaluation.} 
    \label{alg:safecheck}
    \begin{algorithmic}
         \STATE Simulate $I_i$ taking the action $a_t$
         \STATE Get Kinematic predictions from HPN, i.e., $\Tilde{\textbf{K}}_{t+1:t+M}^{\prime}$ = HPN ($\Tilde{\textbf{K}}_{t-N:t}$) for all vehicles in the road $c \in \mathcal{C} = (\widetilde{\mathcal{I}} \cup  \widetilde{\mathcal{H}})$
         \FOR{$t = t+1$ to $t+M$ (Compute safety score for $M_{steps}$ predictions)} 
         \STATE Compute $ttc_t$ between $I_i$  and all $c \in \mathcal{C}   \setminus \{I_i\}$ using $x,y,v,\psi$ at time $t$
          \STATE Compute $min(ttc_t)$
          \STATE Get next prediction at $t = t+1$ 
         \ENDFOR   
         \STATE Compute $safe_{score}$ using the predicted $ttc_{t+1:t+M}$ 
         \STATE $safe_{score} = \frac{\sum_{i=t+1}^{t+M} w_{i}ttc_{i} }{\sum_{i=t+1}^{t+M} w_{i}}$ 
         
        
         
         \IF{$safe_{score}$ $ <  safe_{th}$ or $any(ttc_{t+1:t+M})<  critical_{th}$}
         \STATE Return unsafe
         \ELSE
          \STATE Return safe
         \ENDIF
        
    \end{algorithmic}
\end{algorithm}

\begin{algorithm}[t]
    \caption{Action selection.} 
    \label{alg:safeaction}
    \begin{algorithmic}
        \STATE Initialize $\widetilde{\mathcal{A}}_{safe}$ = $\mathcal{A}$
        \WHILE{$\widetilde{\mathcal{A}}_{safe}$ is not empty}
        \IF{during training}
        \STATE Select $a_t$ following the exploration policy on set $\widetilde{\mathcal{A}}_{safe}$ 
        \ELSIF{during test}
            \STATE Select $a_t = \max_{a' \in \widetilde{\mathcal{A}}_{safe} } Q(s_{t},a';\textbf{w})$ 
        \ENDIF
        \IF{$a_t$ is safe (Algorithm \ref{alg:safecheck})}
        \STATE Return $a_t$ 
        \ELSE
        \STATE Remove $a_t$ from $\widetilde{\mathcal{A}}_{safe}$
        \ENDIF
     \ENDWHILE
     \STATE Compute the $safety_{score}$ as in Algorithm~\ref{alg:safecheck}
     \STATE Return $a_t$ with highest $safety_{score}$ in $\mathcal{A}$
    \end{algorithmic}
\end{algorithm}

\subsubsection{Safe Value Function Network (VFN)}
The VFN estimates the state-action value function. The combination of prediction (HPN) and decision-making (VFN) allows prediction-aware planning and improves the AVs' ability to learn to navigate complex scenarios, and the safety prioritizer further increases safety. The proposed approach utilizes deep reinforcement learning (DRL) to achieve a high-level policy for safe tactical decision-making. 
As presented, the input consists of a stack of $N$ past observations and $M$ future hypotheses, i.e., $s_t = [\Tilde{\textbf{o}}_{t-N:t} ,\Tilde{\textbf{o}}_{t+1:t+M}]$, and the 3D CNN operates as a feature extractor. The VFN is trained to learn the optimal Q-values that maximize our social reward function, optimizing social utility.
During training, agents are trained in a semi-sequential manner, as in~\cite{toghi2021altruistic}. 

The VFN outputs the Q-values that are masked by a safety prioritizer, constraining the RL policy to a safe action space. Therefore, in our framework, when the agent policy chooses an unsafe action, the safety prioritizer masks the action and selects a safer action, saving the unsafe action ($a_t$) and the associated state in the $RM$ with a negative reward ($r_{unsafe}$). By reducing episode restarts due to potential collisions, the safety prioritizer increases sample efficiency and safety.

The proposed prediction-aware planning and the social-aware optimization algorithm is described in \textbf{Algorithm}~\ref{alg:Joint_Pred_algorithm}. We first run a batch of sample simulations to pre-fill our replay buffer before starting the learning phase. To account for the unbalance in training data, the experience replay buffer is re-balanced~\cite{toghi2021social}.

\begin{algorithm}[t]
    \caption{Prediction-aware planning DDQN.} 
    \label{alg:Joint_Pred_algorithm}
    \begin{algorithmic}
         \STATE Define and Initialize \textit{Replay Memory buffer} $RM$.
         \STATE Define and Initialize action-value function $\Tilde{Q}(.;\textbf{w}^-)$  and  target network $\Tilde{Q}(.;\hat{\textbf{w}})$ with $\textbf{w}^-=\textbf{w}_{ini}$ and $\hat{\textbf{w}}=\textbf{w}^-$
          \STATE Save in the $RM$ the first's $E_{ini}$ episodes. 

         \FOR{$\mathrm{e}=E_{ini}$ to $N_{\mathrm{episode}}$} 
         \STATE Obtain observation history $\Tilde{\textbf{o}}_{t-N:t}$
         \STATE Predict $M$ hypothesis $\Tilde{\textbf{o}}_{t+1:t+M}^{\prime}$ (\textbf{Algorithm}~\ref{alg:hpn})
         \STATE Compute $s_t = [\Tilde{\textbf{o}}_{t-N:t} ,\Tilde{\textbf{o}}_{t+1:t+M}^{\prime}]$
              \FOR{$t = t_{ini}$ to T }
                \FOR{$I_i$ in $\mathcal{I}$}
                    \STATE For agents $I_j$, $j \neq i$, freeze $\textbf{w}^-$ 
                     \FOR{$N_{iterations}$}
                         \STATE With probability $\epsilon$ choose $a_t$ randomly,
                         \STATE else choose $a_t = \max_{a' \in A} Q(s_t,a';\textbf{w}^+)$
                         \STATE Verify action $a_t$ (\textbf{Algorithm}~\ref{alg:safecheck})
                         \IF {$a_t$ is not safe} 
                          \STATE Store transition $(s_t, a_t , r_{unsafe} , \emptyset$) in $RM$
                         \STATE $a_t$ = Select a safe action (\textbf{Algorithm}~\ref{alg:safeaction}) 
                         \ENDIF
                          \STATE Take $a_t$ ($a_{safe}$), and observe $r_t, \Tilde{\textbf{o}}_{t+1}$ 
                         \STATE Store transition $(s_t, a_t , r_t , s_{t+1}) $ in $RM$
                        \STATE Compute $\textbf{w}^+_{k+1} \leftarrow \textbf{w}^+_k - \alpha \hat{\nabla}_\textbf{w} \mathcal{L}(\textbf{w}^+)$

              \ENDFOR
              \STATE Disseminate weights $\textbf{w}^- = \textbf{w}^+$ for all $I_i \in \mathcal{I}$
            \ENDFOR
            \STATE Reset $\hat{\textbf{w}} \leftarrow \textbf{w}^-$ every $\quad Target_{update} \quad$
          \ENDFOR
        \ENDFOR
    \end{algorithmic}
\end{algorithm}

%
\section{Experimental Results}
\label{sec:experimentalresults}
\noindent This section begins with a description of the simulation environment and the HV model in mixed-autonomy traffic. Before presenting our findings, practical aspects of training and validation are explored. Finally, we present our results showing the importance of prediction-aware planning for cooperative driving.
\subsection{Implementation Details} 
\label{sec:implementation}
\subsubsection{RL environment and Computational Details}
\label{sec:drivingsimulator}
We customize the OpenAI Gym Highway environment in~\cite{leurent2019approximate}. We design five scenarios for our experiments, i.e, a straight highway, highway exiting, highway merging, intersection, and roundabout scenarios ($f_h, f_e, f_m, f_i, f_r \in \mathcal{F} $). The AVs are trained surrounded by HVs with various behaviors, i.e, conservative, moderate, and aggressive, ($b_c,b_m,b_a \in \mathcal{B}$). A scenario with mixed behavior is obtained by sampling from the behaviors in $\mathcal{B}$ for each HV. 
The VFN is trained for $N_{episodes} = 15,000$ episodes and multiple iterations of the training procedure are carried out to guarantee the convergence of the policies. Table~\ref{table: hyperparameters} lists our training and simulation parameters.

\vspace{10pt}
\begin{table}[t]
\caption{\small{Simulation parameters.}}
\begin{center}
\begin{tabular}{c c | c c}
Parameter &
Value &
Parameter &
Value\\ 
\hline
\hline
K prediction &
GP, RBF kernel &
$N_{\mathrm{episode}}$ &
15,000 \\

Prediction Horizon &
4s &
$\epsilon$ decay &
Linear  \\

History window &
2s &
$RM$ buffer size &
8,000  \\ 

Latent Dimension &
512&
Initial exploration $\epsilon_0$ &
1.0 \\ 
Batch size & 
64 &

Final exploration & 
0.05 \\ 
Learning rate $\alpha_0$ &
0.0005 &
Optimizer &
ADAM \\ 
$Target_{update}$ &
300 &
Discount factor $\gamma$ &
0.95 \\ 
\hline
\end{tabular}
\end{center}
\label{table: hyperparameters}
\end{table}
\subsubsection{Driver Modeling}
\label{sec:humandrivermodel}
We model the HV's lateral behavior using the MOBIL model~\cite{kesting2007general} and the longitudinal behavior of HVs is based on the \textit{Intelligent Driver Model} (IDM)~\cite{treiber2000congested}. MOBIL is based on the safety and incentive criteria. For safety, it verifies $\mathrm{a}^{}_n>-b_{\mathrm{safe}}$, where $\mathrm{a}^{}_n$ is the deceleration of the following vehicle in the new lane, and $-b_{\mathrm{safe}}$ is a safe threshold.  Then MOBIL verifies the incentive to change lane measured by $\mathrm{a}'_{ego}-\mathrm{a}_{ego}+ \mathrm{p}\Big( (\mathrm{a}'_n-\mathrm{a}^{}_n) + (\mathrm{a}'_o-\mathrm{a}^{}_o) \Big) > \Delta a_{th}$, where $\mathrm{p}$ is the politeness term and $\mathrm{a}^{}_{ego}$ , $\mathrm{a}^{}_{n}$ and $\mathrm{a}^{}_{o}$ are the accelerations of the ego-HV, the new following vehicle, and the old following vehicle, respectively. Finally, based on the MOBIL model, if both criteria are verified, then the HV performs a lane change.

The longitudinal behavior of HV $k$ is modeled by using the IDM model which computes the acceleration $\dot{v}_{\mathrm{k}}$ as $\dot{v}_{\mathrm{k}}=\mathrm{a}_\mathrm{max}\Big[ 1- \Big( \frac{v_k}{v_{\mathrm{k}}^0} \Big)^\delta - \Big( \frac{d^*(v_k, \Delta v_k)}{d_k} \Big)^2 \Big]$; in which $\delta$ is the exponent of acceleration,  $d_k$ is the gap, $v_{\mathrm{0}}^k$  is the preferred speed, and $v_k$ is the current speed. Following, the preferred minimum gap is computed as $d^*(v_k, \Delta v_k) = d_k^0 +v_kT_\mathrm{k}^0 + \frac{v_k \Delta v_k}{ (2\sqrt{\mathrm{a}_{\mathrm{max}}.\mathrm{a}_{\mathrm{des}}})}$, where $d_k^0$ is the minimum distance, $T_k^0$ is the safe time gap, $\mathrm{a}_{\mathrm{max}}$ is the acceleration limit, and $\mathrm{a}_{\mathrm{des}}$ is the deceleration limit.

We use the centrality metrics as in~\cite{valiente2022robustness,chandra2020cmetric} to obtain the parameters $\mathcal{P}$ that simulate the driver behaviors for the MOBIL and IDM models. 
In our scenarios, the computed parameters $\mathcal{P}$ that represent aggressive, conservative, and moderate behavior are shown in Table~\ref{table:parameters}.

\begin{table}[htbp]
\caption{\small{Computed parameters $\mathcal{P}$ for different simulated driver behaviors.}}
\centering
\begin{tabular}{c|c|ccc} 

Model & Parameter & Aggressive  &  Moderate & Conservative \\
\hline
\hline
MOBIL & $\mathrm{p}$ & 0  & 0.3 & 1\\ 
& $\Delta a_{th}$ & 0 $m/s^2$ & 0.1 $m/s^2$ & 0.4 $m/s^2$ \\ 
& $b_{\mathrm{safe}}$ & 12.0 $m/s^2$ & 6.0 $m/s^2$ & 2.0 $m/s^2$\\ 

\hline
IDM & $T^0$ & 0.5s   & 1s & 3s \\ 
& $d^0$ & 1 $m$ & 2 $m$ & 6.0 $m$\\ 
& $\mathrm{acc}_{\mathrm{max}}$  & 7.0 $m/s^2$ & 3.0 $m/s^2$ & 1.0 $m/s^2$\\ 
& $\mathrm{acc}_{\mathrm{des}}$ & 12.0 $m/s^2$        & 7.0 $m/s^2$ & 2.0 $m/s^2$\\ 
\hline
\end{tabular}

\label{table:parameters}
\vspace{-10pt}
\end{table}

\subsection{Evaluation Metrics and Hypotheses}
\label{sec:performancemeasures}

\noindent The system's performance is evaluated based on safety, effectiveness, and prediction error. We select two indicators that, notwithstanding their correlation, offer distinct perspectives on the effectiveness of our approach. We calculate the proportion of episodes with at least one crash ($C(\%)$) in order to measure safety. The vehicles' average traveled distance ($DT(m)$) is utilized to measure efficiency. Finally, the prediction error is measured in terms of the prediction reconstruction error ($PRE$) of the $VM$ predictions and in terms of position error ($PE$) for the Kinematic prediction $K$. Based on those evaluation metrics we investigate the following hypotheses:

\begin{itemize}
    \item \textbf{H1.} \emph{The GP is a more powerful approach to predicting time series when compared to the AE for Kinematic predictions, therefore, using the GP for kinematic prediction improved the prediction performance when measured by the position error $PE$. Additionally, temporal information is important for accurate prediction, therefore we expect a higher performance of the $VM$ prediction from our predictive autoencoder when using the observation history, measured by the prediction reconstruction error ($PRE$).}
    \item \textbf{H2.} \emph{The ability to forecast future states improve decision-making in AVs, therefore we anticipate a performance improvement of our prediction-aware VFN measured in terms of safety and efficiency when using the HPN.}
\end{itemize}

\subsection{Analysis and Results}
\begin{figure}[t!]
  \centering
  \includegraphics[width=.48\textwidth]{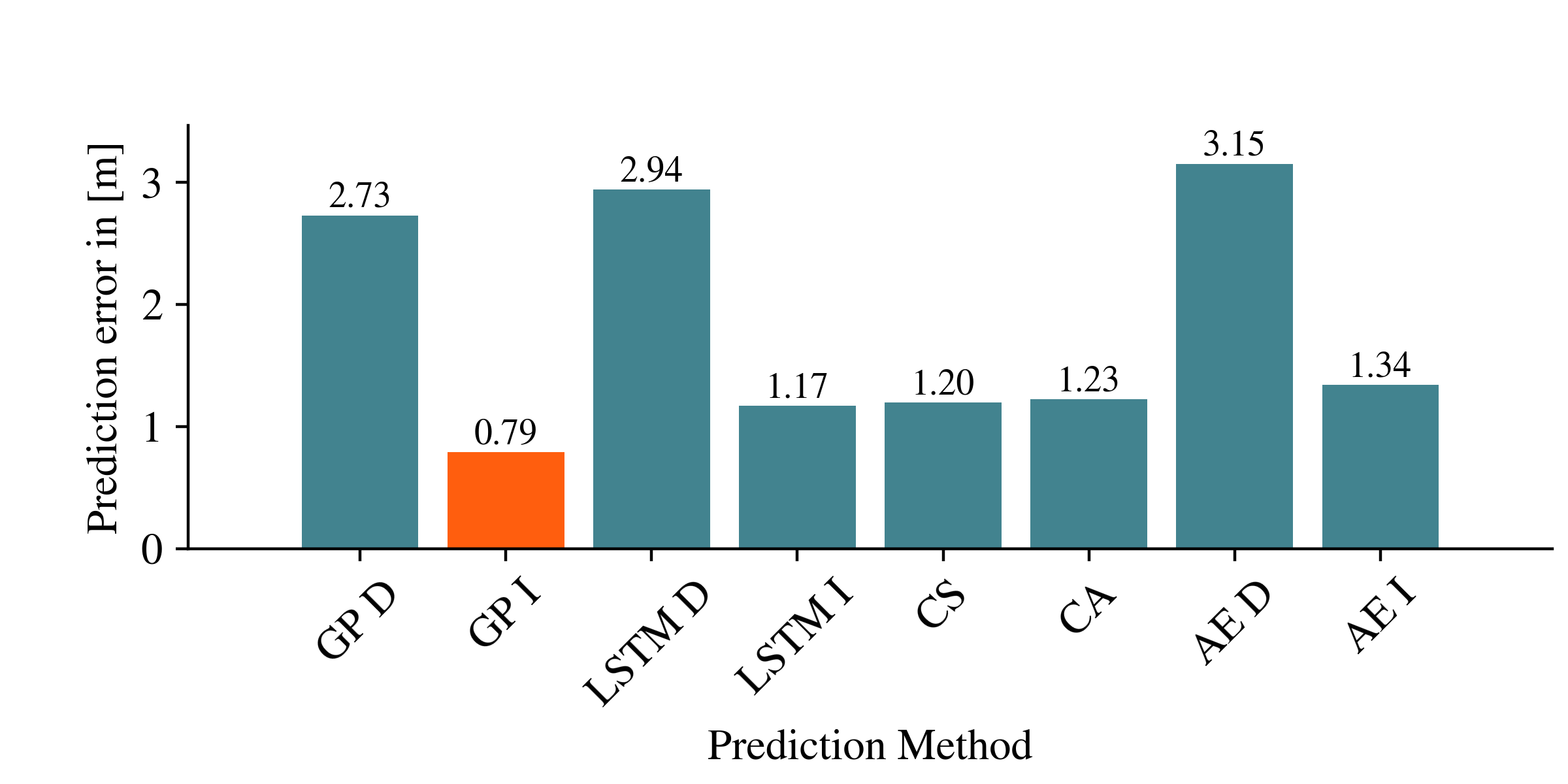}
  \caption{\small{Kinematic prediction baseline comparison in terms of position error ($PE$) in meters.}}
    \label{fig:Prediction_methods}
\end{figure}
\begin{figure}[b!]
  \centering
  \includegraphics[width=.48\textwidth]{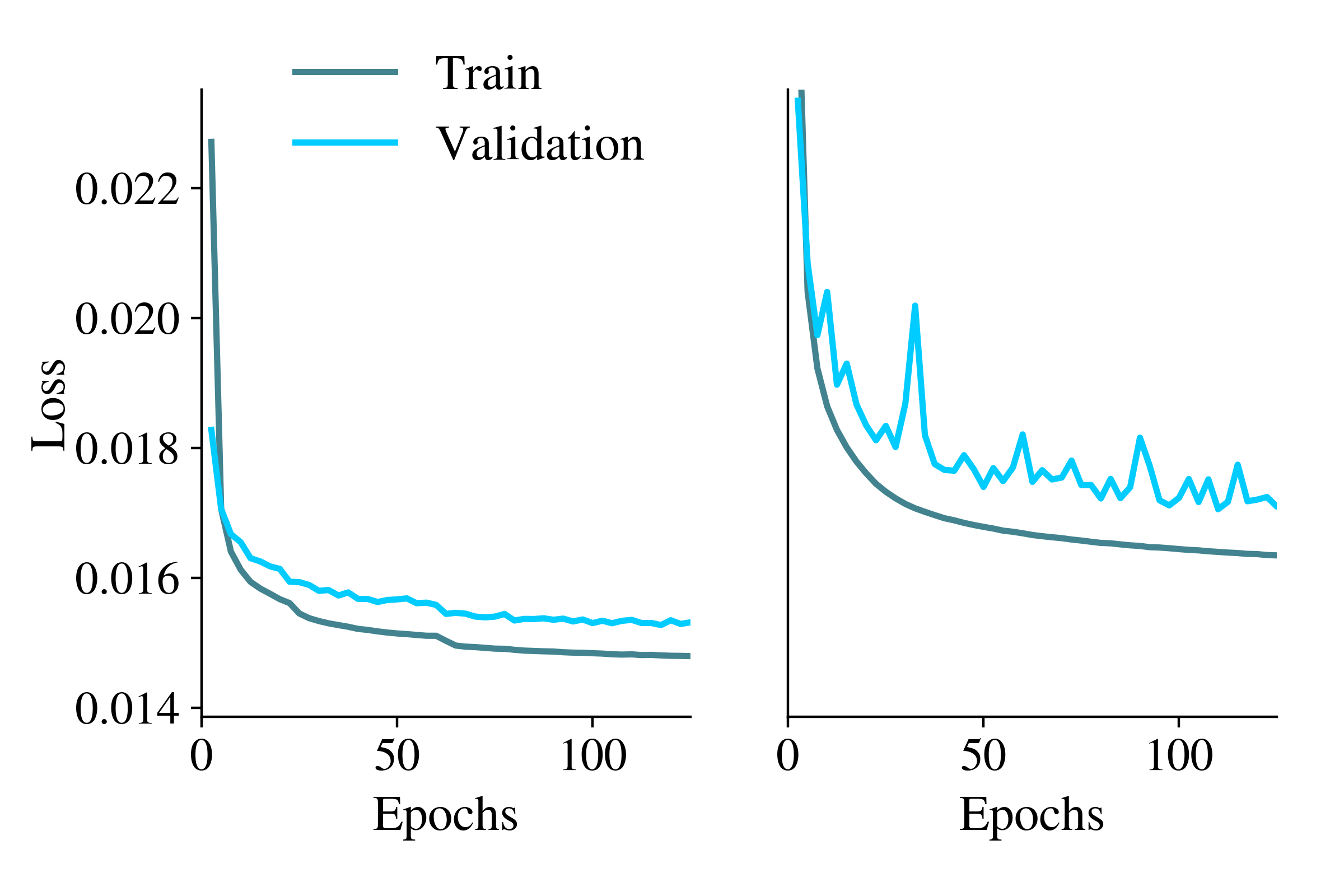}
  \caption{\small{Training and validation loss of the predictive network using observation history (left), and without history (right).}}
\label{fig:ae_loss}
\end{figure}
\begin{figure*}[t!]
  \centering
  \includegraphics[width=.98\textwidth]{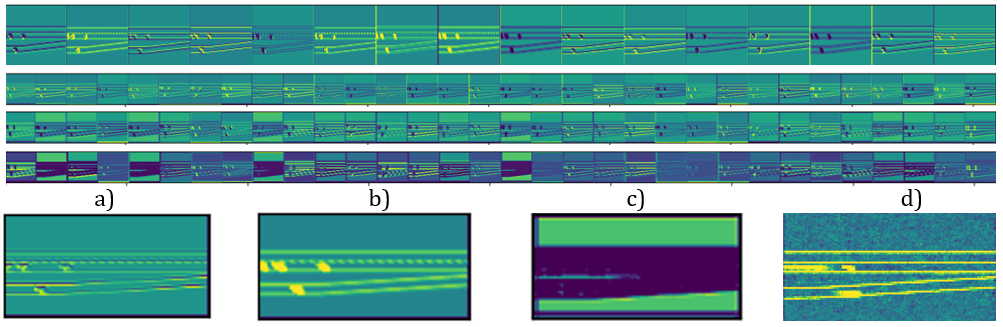}
  \caption{\small{Internal representations of the features at different layers for a merging scenario (Top), selected Internal representations that have been zoomed in (Bottom).}}
\label{fig:pn_features}
\end{figure*}

\begin{figure}[t!]
  \centering
  \includegraphics[width=.48\textwidth]{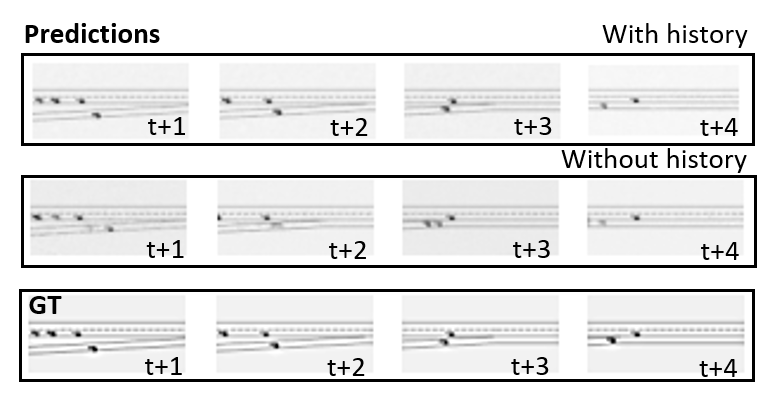}
  \caption{\small{Prediction chain for merging when using observation history (top), and without history (bottom).}}
  \label{fig:chain_merging_prediction}
\end{figure}

\subsubsection{Learning how to predict, Hybrid predictive network (HPN)}
Predicting the actions of HVs is a crucial component of AVs' decision-making. We take advantage of this feature and investigate how incorporating prediction into our framework improves safety and efficiency. We look into this insight while investigating \textbf{H1}. Particularly, we show that prediction in the image domain allows learning powerful representations, and we present how the HPN learns to predict the $VM$ image and the advantages of using the GP approach to improve Kinematic prediction.

\textbf{Kinematic prediction.}
We first investigate our \textbf{H1} and show how using the GP for kinematic prediction improves the prediction performance. We compare different kinematic prediction baselines and measure their performance using the position error $PE$ in meters. We compare five prediction approaches, i.e, the GP prediction approach, an LSTM network, Constant Speed (CS), Constant Acceleration (CA) based prediction, and the predictive AE kinematic prediction. GP, LSTM, and AE can be used to predict any time series and leveraging that, we consider two approaches for each method: direct and indirect prediction. Therefore, we compare eight baselines the GP direct (GP D), the GP indirect (GP I), LSTM Direct (LSTM D), LSTM Indirect (LSTM I), CS, CA, AE direct (AE D) and AE indirect (AE I).

In a direct prediction approach, $(x,y)$ are regressed by two distinct models learned from the history ($x_{t-N:t},y_{t-N:t}$), producing direct predictions of futures $(x,y)$. Differently, in an indirect prediction approach, the vehicle's heading ($\phi_{t-N:t}$) and speed ($v_{t-N:t}$) histories are considered and the models for heading and speed $(\phi, v)$ are learned using the predictive approach. Using the learned models, the predictions for future heading and speed $(\phi, v)$ are computed and utilized to calculate future position ($x_{t+1:t+M},y_{t+1:t+M}$). In our experiments, a compound kernel of linear and RBF is used following previous work~\cite{mahjoub2019representing}.

As illustrated in Figure~\ref{fig:Prediction_methods} the indirect GP (GP I, in orange) approach outperforms the other baselines in terms of $PE$, showing how this non-parametric Bayesian scheme allows the incorporation of complex model structures and is a suitable option for our kinematic prediction method, verifying our \textbf{H1}.

\textbf{Observation history.}
Together with the Kinematic prediction, the HPN outputs the velocity map image prediction ($VM$). A prediction reconstruction error ($PRE$) loss is utilized to calculate the error between the predicted observation $o^{pred}$ and the corresponding $o$, i.e., $L_2(\boldsymbol{o_{t+1}}, \boldsymbol{o_{t+1}^{\prime}}) = \sum_i (o_i-o_i^{\prime})^2$.
We evaluate the HPN's performance using only the current observation as input and history of $N$ observations, demonstrating the importance of temporal information for accurate prediction measured by the $PRE$. 
Figure~\ref{fig:ae_loss} depicts the training loss results, where the left image is for the HPN that uses the history of $N$ observation and the right image uses just the current observation. As shown in the figure, when using the temporal information the $PRE$ loss is approximately $20\%$ smaller when using history (left) than without history (right). Similarly, Figure~\ref{fig:chain_merging_prediction} shows the qualitative output of the HPN prediction chain using the current observation or a history of observations as input. The results with history (top) show clearer and more accurate visual predictions and confirm why the $PRE$ is lower when using the history. 

Figure~\ref{fig:pn_features} presents some qualitative results of the internal representations at different layers of the HPN for a merging scenario. In Figure~\ref{fig:pn_features} (bottom), a zoomed-in version of some internal representations are illustrated for visualization. As observed, the HPN learns to extract and highlight important information from the input observation, such as (a) lanes, (b) road agents, (c) road segments, and (d) possible hypotheses on how the environment evolves. Despite the fact that the HPN has not been trained for a segmentation task, it learns to segment the road, agents, and lanes, which could be useful information for the prediction and driving tasks.


\subsubsection{Safer VFN leveraging prediction}
\begin{figure}[t!]
  \centering
  \includegraphics[width=.4\textwidth]{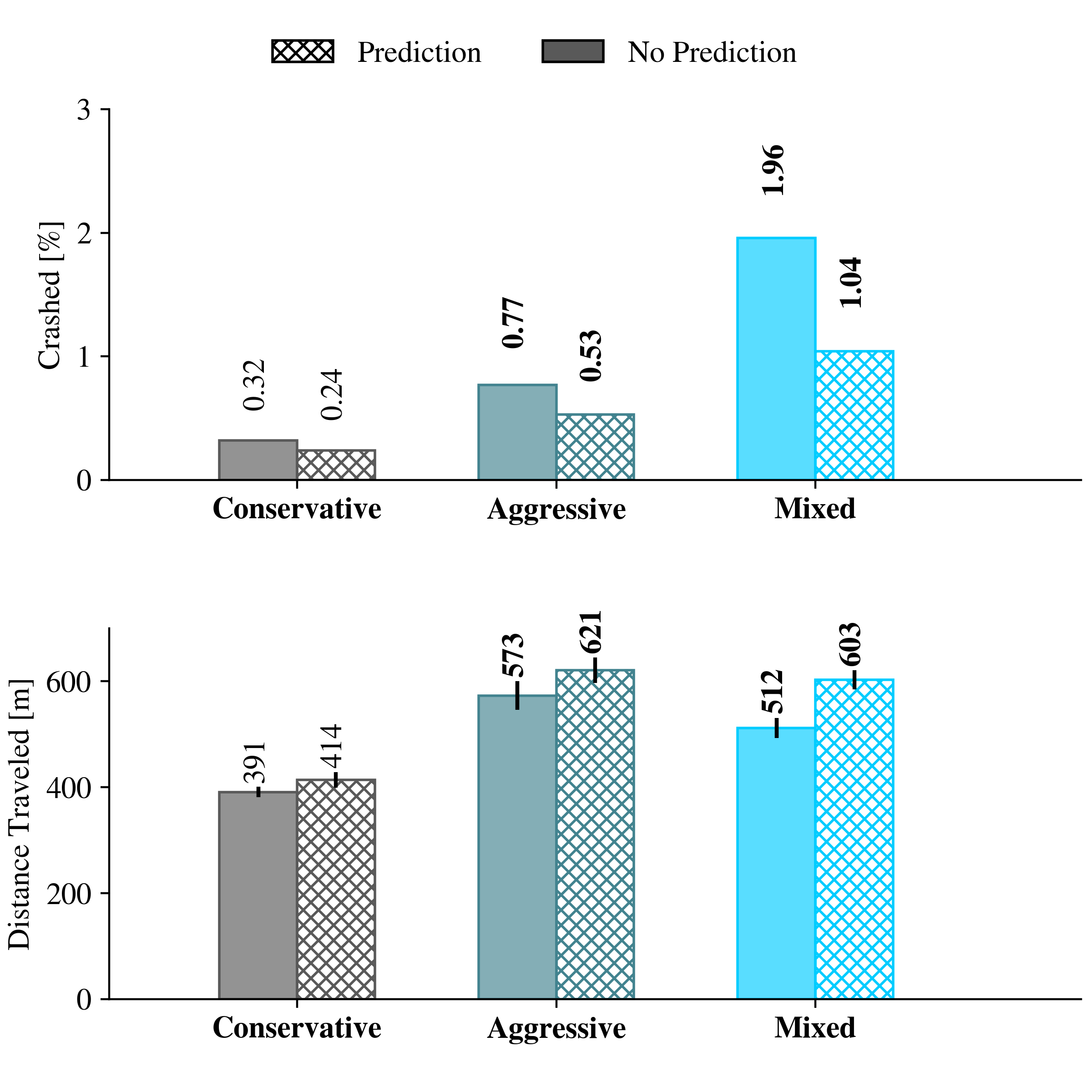}
  \caption{\small{Performance enhancement in a highway merging scenario resulted from using prediction. Safety is measured in terms of crash percentage (Top, $C(\%)$) and efficiency by average traveled distance (Bottom, $DT(m)$)}.}
  \label{fig:comparison}
\end{figure}

Using the HPN and prediction chain, the VFN is trained to optimize for a social utility. We evaluate the performance of the VFN when using just the history as input, i.e, $s_t = [\Tilde{\textbf{o}}_{t-N:t}]$ and when additionally utilizing the prediction output of the prediction chain, i.e. $s_t = [\Tilde{\textbf{o}}_{t-N:t} ,\Tilde{\textbf{o}}_{t+1:t+M}]$. Figure~\ref{fig:comparison} shows performance improvement by using prediction quantified by crash percentage (Top, $C(\%)$) and average distance traveled (Bottom, $DT(m)$). The results present the AV's performance in the highway merging scenario $f_m$, in the presence of HVs with conservative, aggressive or mixed behavior. 
We observe that when train and test are performed in a conservative environment, or in other words, when HV yields and takes safer actions, the gains from prediction capabilities are not as noticeable, whereas, in an aggressive and mixed environment in which the behavior changes, the performance increases are significant. We believe that anticipating the future is especially useful in those scenarios, which is why performance has improved.

We evaluate the effectiveness of our architecture in diverse scenarios, as well as the performance enhancement of leveraging prediction. We argue that by using prediction, we provide the VFN a prior on how the world will evolve which is helpful for decision-making. Table~\ref{table:comparison} presents the results in different traffic scenarios, i.e, Exiting ($f_e$), Merging ($f_m$), Roundabout ($f_r$), intersection ($f_i$), and Highway ($f_h$) under mixed HV behaviors ($b \in \mathcal{B}$). We compare our architecture when using prediction (VFN+P), and without prediction (VFN) with other related architectures~\cite{van2016deep,toghi2021altruistic, toghi2021cooperative}. 
Our architectures as shown in Table~\ref{table:comparison} outperform the alternative methods, and the improvements are particularly pronounced in the more complex scenarios. 

\begin{figure}[t!]
  \centering
  \includegraphics[width=.48\textwidth]{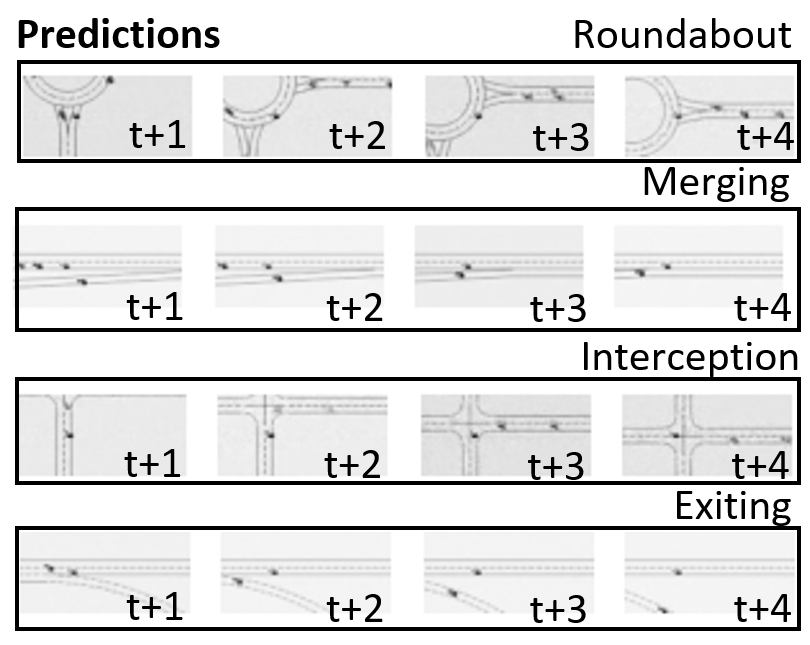}
  \caption{\small{Qualitative results of the prediction chain output for multiple scenarios}}
  \label{fig:chain_multiple}
\end{figure}

The combination of prediction (HPN) and decision-making (VFN) allows for prediction-aware planning and improves the AVs' ability to learn to navigate complex scenarios, and the safety prioritizer is further improved by leveraging the information provided by the prediction chain, increasing safety and efficiency. The results presented in Figure~\ref{fig:comparison} and Table~\ref{table:comparison} verify our \textbf{H2}. Additionally, Figure~\ref{fig:chain_multiple} provides the output of the prediction chain for different traffic scenarios, showing qualitative results that further illustrate the capabilities of the prediction network to forecast the future.

\begin{table}[htbp]
\caption{Performance Comparison (Measured by $C (\%)$) of related architectures showing the performance improvement of our predictive VFN, particularly in challenging scenarios such as intersection and roundabout. The results are shown in different scenarios, Exiting ($f_e$), Merging ($f_m$), Roundabout ($f_r$), intersection ($f_i$), and Highway ($f_h$).}
\centering
\begin{tabular}{c|ccccc} 
 Approach & $f_e$  &   $f_m$     & $f_r$    &   $f_i$   & $f_h$   \\
\hline
\hline
Conv2D+DQN~\cite{van2016deep} & 24.62 &  29.12 &  49.03 &  54.78  & 17.21  \\

Conv3D+A2C~\cite{toghi2021altruistic} & 9.23 &  14.99 &  21.17 &  36.62  & 7.43 \\

Conv3D+DQN~\cite{toghi2021cooperative} & 3.91 &  2.59 &  14.62 &  24.30  & 1.31  \\

Safe DQN~\cite{valiente2022robustness} & 2.51 &  1.95 &  9.04 &  18.67  & 0.44  \\

\textbf{VFN}   &  2.47 &  1.96 &  8.94 & 17.90  & 0.39  \\

\textbf{VFN + P}   & \textbf{1.91}  & \textbf{1.04}  & \textbf{7.01}  &  \textbf{11.10}   & \textbf{0.31}  \\

\hline
\hline
\end{tabular}

\label{table:comparison}
\end{table}

%
\section{Conclusion}
\label{sec:concluding}
\noindent We propose the integration of two crucial components for AVs, social navigation and prediction. The safety and reliability of AVs depend on their predictive capabilities, social awareness, and ability to engage in complex social interactions. For that reason, we propose prediction-aware planning and social-aware optimization in a cooperative RL framework, to allow safe and socially-desirable outcomes. We provide AVs the ability to anticipate the future, allowing them to take informed decisions and proactive actions in AV-HV social interaction scenarios. The safety prioritizer leverages interpretable kinematic predictions from the HPN to restrict the RL policy to assure safe decision-making, reducing future high-risk actions, increasing awareness of the immediate risks, and consequently decreasing crashes. We compare our prediction-aware AV to other solutions and demonstrate how our approach consistently improves safety and efficiency on the road in multiple scenarios.

Further research needs to be done using real human driver data, and more complex traffic scenarios. We plan to extend this work in this direction and show robust generalization capabilities of our agents. We intend to explore learning meaningful and interpretable representations and predictions to help build intuition on the AV's decision-making process.

\ifCLASSOPTIONcaptionsoff
 \newpage
\fi

\bibliographystyle{IEEEtran}
\bibliography{IEEEbibs}

\vskip -2.5\baselineskip plus -1fil
\begin{IEEEbiography}[{\includegraphics[width=1in,height=1.25in,clip,keepaspectratio]{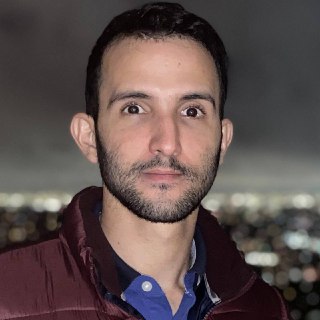}}]{Rodolfo Valiente}
is a Ph.D. candidate in Computer Engineering at the University of Central Florida. His research interests include connected autonomous vehicles, reinforcement learning, computer vision, and deep learning with a focus on the autonomous driving problem. He received a M.Sc. degree from the University of Sao Paulo (USP) in 2017 and his B.Sc. degree from the Technological University Jose Antonio Echeverria in 2014.
\end{IEEEbiography}
\vskip -2.5\baselineskip plus -1fil
\begin{IEEEbiography}[{\includegraphics[width=1in,height=1.25in,clip,keepaspectratio]{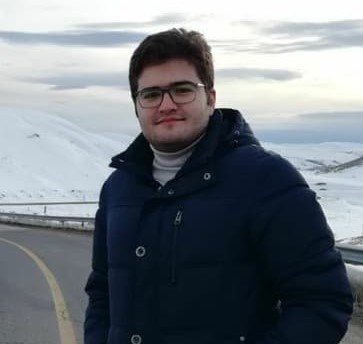}}]{Mahdi Razzaghpour} is a Ph.D. candidate in Computer Engineering at the University of Central Florida and a member of the Connected and Autonomous Vehicles Research Lab (CAVREL). His research interests include Reinforcement Learning, Machine Learning, and deep learning with a focus on the Cooperative driving problem. He received the M.Sc. degree in Computer Engineering from the University of Central Florida in 2021 and the B.Sc. degree in Electrical Engineering from Sharif University of Technology in 2019.
\end{IEEEbiography}
\vskip -2.5\baselineskip plus -1fil
\begin{IEEEbiography}[{\includegraphics[width=1in,height=1.25in,clip,keepaspectratio]{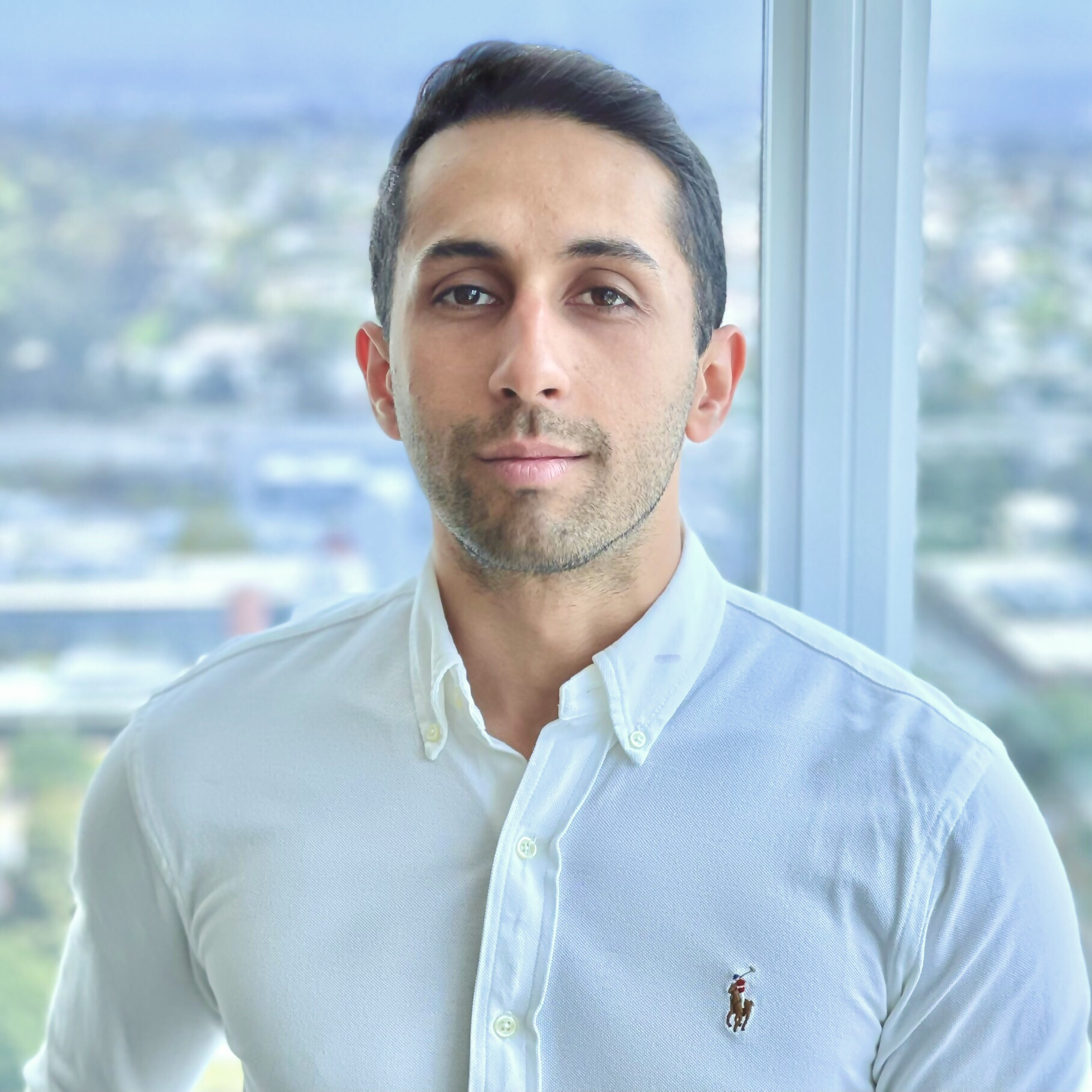}}]{Behrad Toghi}
is a Ph.D. candidate at the University of Central Florida. He received the B.Sc. degree in electrical engineering from Sharif University of Technology in 2016 and has worked as a research intern at Mercedes-Benz R\&D North America and Ford Motor Company R\&D between 2018 and 2020. His work is in the intersection of artificial intelligence and cooperative networked systems with a focus on autonomous driving.
\end{IEEEbiography}
\vskip -2.5\baselineskip plus -1fil
\begin{IEEEbiography}[{\includegraphics[width=1in,height=1.25in,clip,keepaspectratio]{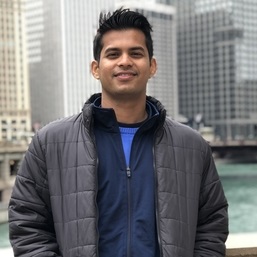}}]{Ghayoor Shah}
is a Ph.D. candidate at the University of Central Florida. He received the B.Sc. degree in Computer Engineering from University of Illinois at Urbana-Champaign in 2018. He has previously worked as a mobility engineering intern at Phantom Auto and as a research intern at Ford Motor Company. 
His research interests include Connected and Autonomous Vehicles (CAVs), scalability analysis of V2X, and applications of artificial intelligence to cooperative driving. 
\end{IEEEbiography}
\vskip -2.5\baselineskip plus -1fil
\begin{IEEEbiography}[{\includegraphics[width=1in,height=1.25in,clip,keepaspectratio]{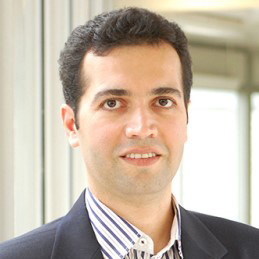}}]{Yaser P. Fallah} is an Associate Professor in the ECE Department at the University of Central Florida. He received the Ph.D. degree from the University of British Columbia, Vancouver, BC, Canada, in 2007. From 2008 to 2011, he was a Research Scientist with the Institute of Transportation Studies,
University of California Berkeley, Berkeley, CA, USA. His research, sponsored by industry, USDoT, and NSF, is focused on intelligent transportation systems and automated and networked vehicle safety systems.
\end{IEEEbiography}

\end{document}